\documentclass[runningheads,a4paper]{llncs}

\usepackage{amssymb}

\usepackage{arydshln}
\usepackage[pdftex]{graphicx}

\usepackage[breaklinks,colorlinks=true,citecolor=blue]{hyperref}

\begin{document}

\mainmatter  

\title{Gliders2012: Development and Competition Results}

\titlerunning{Development and Competition Results}

%
%
\author{Edward Moore\inst{1}\thanks{CSIRO authors are listed in alphabetical order.} \and Oliver Obst\inst{1} \and Mikhail Prokopenko\inst{1} \and Peter Wang\inst{1}\\  Jason Held\inst{2}}
\authorrunning{Gliders2012} 
%

\institute{CSIRO Information and Communication Technologies Centre, Adaptive Systems,\\ 
PO Box 76, Epping, NSW 1710, Australia
\and
Saber Astronautics Australia, 53 Balfour St, Chippendale, NSW 2008, Australia}


\maketitle
\setcounter{footnote}{0}

\begin{abstract}
The RoboCup 2D Simulation League incorporates several challenging features, setting a benchmark for Artificial Intelligence (AI). In this paper we describe some of the ideas and tools around the development of our team, Gliders2012. In our description, we focus on the evaluation function as one of our central mechanisms for action selection. We also point to a new framework for watching log files in a web browser that we release for use and further development by the RoboCup community. Finally, we also summarize results of the group and final matches we played during RoboCup 2012, with Gliders2012 finishing 4th out of 19 teams.
\end{abstract}

\section{Introduction}

The RoboCup Simulation League \cite{Kitano97} incorporates several challenging features, setting a benchmark for Artificial Intelligence (AI). 
The following list includes some of the most prominent characteristics of the RoboCup 2D Simulation League:
\begin{itemize}
\item distributed  client/server  system  running  on  a  network, leading to fragmented,  localized  and  imprecise  (noisy  and latent)  information  about  the  environment  (field) \cite{Noda2003};
\item concurrent  communication  with  a  medium-sized   number of  agents \cite{Stone99};
\item heterogeneous  sensory  data  (visual,  auditory,  kinetic) and limited  range  of basic commands/effectors  (turn,  kick,  dash, $\ldots$)  \cite{ATAL2000};
\item asynchronous  perception-action  activity and limited  window of opportunity  to  perform  an  action \cite{Butler2001};
\item autonomous decision-making under  constraints  enforced  by  teamwork (collaboration)  and opponent  (competition) \cite{IAAI2000};
\item conflicts  between reactivity  and deliberation \cite{Reis2001}; 
\item no  centralized  controllers and centralized world model (no global  vision, etc.) \cite{Prokopenko02,Prokopenko03}.
\end{itemize}

From the onset of the RoboCup effort it was recognized that, as a benchmark, RoboCup is fairly different from another classical AI problem --- chess. As pointed out by Asada et al. \cite{Asada99}, chess and RoboCup differ in a few key elements: environment (static vs dynamic), state change (turn-taking vs real-time), information accessibility (complete vs incomplete),
sensor readings (symbolic vs non-symbolic), and control (central vs distributed). This difference has been well understood over the last decade. Nevertheless, there are some similarities, for example, efficient evaluation functions used by the RoboCup agents are conceptually similar to evaluation functions used by chess computers: in either case the agent is attempting to consider multiple future states, assign some values to the alternative outcomes, and choose an action optimizing the evaluations. One may argue that superior performance of recent  world champions  in the RoboCup 2D Simulation League \cite{Akiyama2010,Bai2011} may be attributed, at least partially, to sophisticated evaluation functions employed by these teams. In this short paper we describe a novel mechanism utilizing action-dependent evaluation functions, comparing it to some well known constructive models used by belief revision and belief update \cite{Peppas96}.

The experiments are carried out using a new simulated soccer team for the RoboCup soccer 2D simulator \cite{Manual2002}, Gliders2012.  The team code is written by C++ using agent2d: the well-known base code developed by Akiyama et al. \cite{agent2d}. Other software packages are used as well:
\begin{itemize}
\item librcsc: a base library for RCSS with various utilities describing relevant geometrical constructs, world model, etc.;
\item soccerwindow2: a viewer program for RCSS, working as a monitor client, a log player and a visual debugger;
\item fedit2: a formation editor for agent2d, allowing to design a team formation.
\end{itemize}

\section{Motivation and approach}

\subsection{Chess analogy}

As argued by Laram{\'e}e, in chess ``the evaluation function, is unique in a very real sense: while search techniques are pretty much universal and move generation can be deducted from a game's rules and no more, evaluation requires a deep and thorough analysis of strategy'' \cite{Chess00}. He lists several main board evaluation metrics: material balance (an account of which pieces are on the board for each side), mobility (a measure of how many move options are available, especially for powerful chess pieces), board control (a side controls a square if it has more pieces attacking it than the opponent), development (minor pieces should be brought into the game as quickly as possible), pawn formations, king safety and tropism (a measure of how easy it is for a piece to attack the opposing king; usually measured in terms of distance).

One may draw some parallels with RoboCup Simulation. For example, the goal safety and distances to the opposing goal are analogous to king safety and tropism, pawn formations may give some hints to team formations, development is somewhat similar to developing an attack from within your own half, board control is akin to blocking and marking opponent players (i.e., field control), mobility is achieved by either positioning teammates to receive a pass, or creating multiple directions for dribble, and material balance can be computed by accounting for heterogeneous player types and remaining stamina values. All these analogies are, of course, not direct --- nevertheless, they may be provide some inspiration for an evaluation function relevant for RoboCup Simulation.

\subsection{Basic evaluation}

The evaluation function of agent2D is, however, quite simple. Using the chess analogy, it implements \emph{tropism} only, and is intended to make the basic client play in a goal-oriented fashion. For a player controlling the ball, it considers two features of each possible resultant state $s$: its $X$-coordinate (the larger the better) and the distance from it to the opponent's goal (the smaller the better). That is, the opponent's goal is the ultimate desirable resultant state $S$, and each action $a$ is rated in terms of a single distance metric $D$
\begin{equation} \label{single} r(a) = D(s = result(a), S) \ . \end{equation}
The action that is selected is simply the one that minimizes the distance between resultant and desirable states, i.e., minimizes this metric:
\begin{equation} \label{argmin} a^* = \arg\min_a r(a) \ . \end{equation}
For the players who are not controlling the ball and are engaged in intercept behavior, the evaluation function is not specified explicitly. These players select positions on the field according to their roles in the team formation.

The evaluation function (\ref{argmin}) has reached a significant aim: all types of actions (dribbles, passes, etc.) can be directly compared to each other in terms of a single metric. At the same time, the simple computation is not adequate to support a very sophisticated tactical play in mid-field, or even near opponent's penalty area. Another drawback is that all the actions are judged in relation to a single point: the opponent goal.

\subsection{Multiple desirable states: tactics}

Our main objective was to retain the advantage of a single metric, but diversify the evaluation by considering multiple points as desirable states. Moreover, we suggested~\cite{POWH12} not only that the most desirable  state can change from one cycle to another, but also that a player may entertain multiple desirable  states at any given time (cycle). This diversity is brought about by different tactics. For example, a player may consider one desirable  state $S_1$ if passing to the left (action $a_1$) pursuing one tactic, another resultant state $S_2$ if passing to the right (action $a_2$) guided by another tactic, and yet another desirable state $S_3$ if dribbling to the center (action $a_3$) suggested by a third tactic. Each of the considered actions is evaluated with respect to the corresponding desirable state that represents one of possible tactical ways to develop the play.  

In other words, at any given time, there is a number $m$ of tactics represented by a set of desirable states: $\{S_1, \ldots, S_m\}$, and the feasible actions are partitioned into $m$ sets: $A_1, \ldots, A_m$,  so that for every action $a_i$, there is a set $A_j$ such that $a_i \in A_j$. We denote the function mapping  an action to its tactical state by 
\begin{equation} \label{tactics} tactics: a \rightarrow S \ . \end{equation}
Then each feasible action is rated with respect to the corresponding desirable state:
\begin{equation} \label{mult} r(a) = D(s = result(a), S = tactics(a))  \end{equation}
followed by selection according to the optimization (\ref{argmin}). This approach does not impose tactics in a top-down fashion, selecting one tactic and the sub-selecting the best action for the chosen tactic. Rather, all feasible actions are considered, and tactics contribute to the evaluation via the desirable states suggested by the tactics. 
In certain cases the opponent's goal becomes one of possible desirable states (one of the tactics), keeping the goal-oriented behavior of agents.

The difference between definitions (\ref{single}) and (\ref{mult}) is simply that the desirable states that the player is trying to reach are not independent of actions, but rather \emph{are} action-dependent, and this dependence is tactical. To re-iterate, the comparison between two actions $a_1$ and $a_2$ according to the first definition (\ref{single}) always assumes  the same action-independent state $S$ that is evaluated against, while the proposed definition (\ref{mult}) allows for different desirable states  $S_1 = tactics(a_1)$ and $S_2 = tactics(a_2)$. The metric $D$ is the same for all actions, retaining the advantage of a uniform comparison across different action types. 

We would like to point out at this stage a difference between the proposed action-dependent evaluation function and other action-dependent formalisms, e.g., with action-dependent features generalizing state space proposed by Stone and Veloso \cite{Stone98}. The latter study described a multi-agent learning paradigm called team-partitioned, opaque-transition reinforcement learning (TPOT-RL). TPOT-RL introduced the concept of using action-dependent features to generalize the state space. However, regardless of action-dependent features, each possible action $a$ is evaluated by TPOT-RL based on the current state of the world using a \emph{fixed} function $e: (S, A) \rightarrow U$. That is, the function $e$ is the same for all actions in TPOT-RL. 

Another interesting point is the analogy between multiple desirable states unified by the proposed evaluation function and the constructive model for belief update and belief revision \cite{Peppas96}. Belief revision is the process by which a rational agent changes their beliefs about a static world in the light of new data. Belief update on the other hand is the process by which an agent maintains their beliefs up to date with an evolving world. The constructive model for belief revision includes a single similarity structure centered on all possible worlds consistent with current beliefs (a single system of nested spheres), and identifies the nearest sphere which is consistent with the new data. Peppas et al. \cite{Peppas96} have shown that the model for belief update uses multiple systems of spheres (one for each possible world), finds in parallel the  spheres consistent with the new data that are nearest to their respective central possible worlds, and collects possible worlds within these spheres. Arguably, the action-dependent evaluation proposed here is akin to the constructive model of belief update.

\subsection{Mobility and field control}

The function $tactics$ implements the \emph{mobility} aspect of evaluation, by diversifying options of the player controlling the ball in continuing the game. The other teammates can also use this function in selecting a desirable state for their positioning. That is, a player choosing a position on the field does not have to have a single best point, given the current state. It may consider multiple points, each of which is again dependent on the action. For example, the player may consider state (point) $S_1$ when moving to the left wing with action $a_1$, and  state (point) $S_2$ when blocking a nearest opponent with action $a_2$. Each of the resultant states $s_1 = result(a_1)$ and $s_2 = result(a_2)$ are compared with the corresponding desirable states suggested by the tactics $S_1$ and $S_2$, and the action achieving the best proximity in terms of the  metric $D$ is selected.  
The diversification in positioning achieves both \emph{mobility} (by enabling better passes to these teammates) and \emph{field control} --- by taking key points and blocking key directions.

The idea of field control can be traced to a generic framework describing abstract spatio-temporal relationships  described by Dylla et al. \cite{DFL08}. The latter work did not menion field control explicitly but argued that a reachability relation is needed to express spatial relationships between the players and the ball. They suggested to use Voronoi diagrams: a Voronoi diagram is the partitioning of a plane with $n$ points into $n$ convex polygons such that each polygon contains exactly one point and every point in the given polygon is closer to its central point than any other \cite{DFL08}. This was further developed by Akiyama et al. who used a dual representation of Voronoi diagrams --- the Delaunay triangulation \cite{Akiyama2008,Akiyama2010}. 

\subsection{Example}

\begin{figure}[t]
\centering
\includegraphics[width=9.4cm]{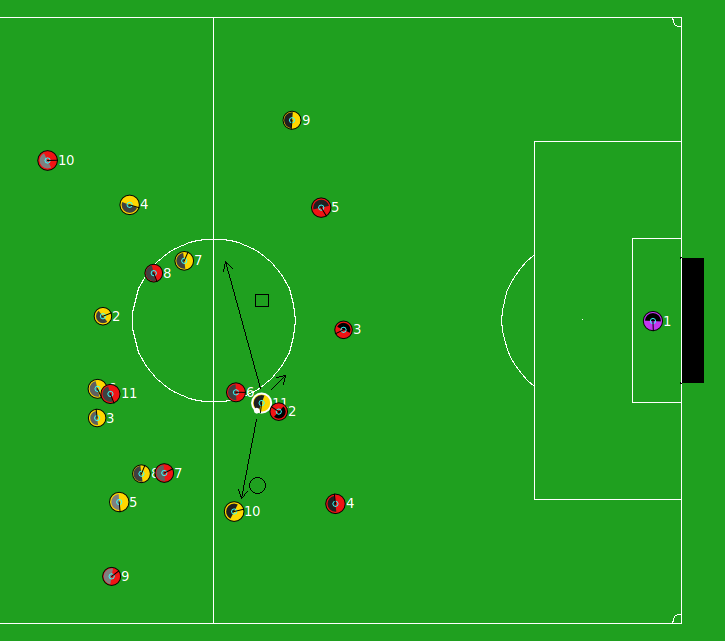}
\caption{Action-dependent evaluation. Small rectangle and circle show different desirable states. The arrows point to possible resultant states. The pass to player 10 is selected since the distance between its resultant and desirable states is the smallest.}
\label{fig1}
\end{figure}

Figure \ref{fig1} illustrates the concept of action-dependent evaluation.  The player controlling the ball (left team, number 11) has several options available: it can dribble in a general forward-left direction, pass to teammates 7 and 10 (in a number of ways, including direct and lead passes), etc.  We consider three choices (shown by arrows): dribble forward-left, pass to the left to teammate 7, and pass to the right to teammate 10. The agent2d's evaluation function would most likely rated the dribble higher, as the resultant state (the arrow-head) has a larger $X$-coordinate and a smaller distance from the opponent's goal than the alternatives. The new evaluation function identifies two desirable states instead, shown by a small rectangle to the left of player 11, and a small circle to its right. The rectangle defines the tactic suggesting to develop an attack to the left and through the center, and the circle corresponds to the tactic preferring the right wing. The dribble and pass to teammate 7 are partitioned to the first tactic (rectangle), and the pass to teammate 10 belongs to the second tactic (circle). Each of these actions is rated by proximity of their resultant states (the arrow-heads) to the rectangle and circle respectively. The pass to number 10 has the smaller distance between the resultant and desirable states, and is then selected.

\section{Development and Results}

The proposed approach was implemented in Gliders2012 --- a new team based on agent2d \cite{agent2d}. We carried out multiple iterative experiments, matching Gliders2012 up against the agent2d (HELIOS Base team), and achieving $\approx +4.0$ goal difference, typically averaged over 100 games. In doing that, some of the most important tools for development were a set of scripts to automate running tournaments. Average results of such a tournament are a useful indicator if a proposed change in the team is really helpful or possibly hurts performance. Oftentimes, it is helpful to briefly watch some of the games that are stored on the simulation server. To make these matches available for viewing over web browser, the current existing solution is to convert them into flash (SWF) format. Even though source code for this program is available~\cite{GO03}, a required library to create SWF files is proprietary and not anymore distributed. As a result, this option is not available for new teams. 

\begin{figure}[htbp]
\centering
\includegraphics[width=0.8\linewidth]{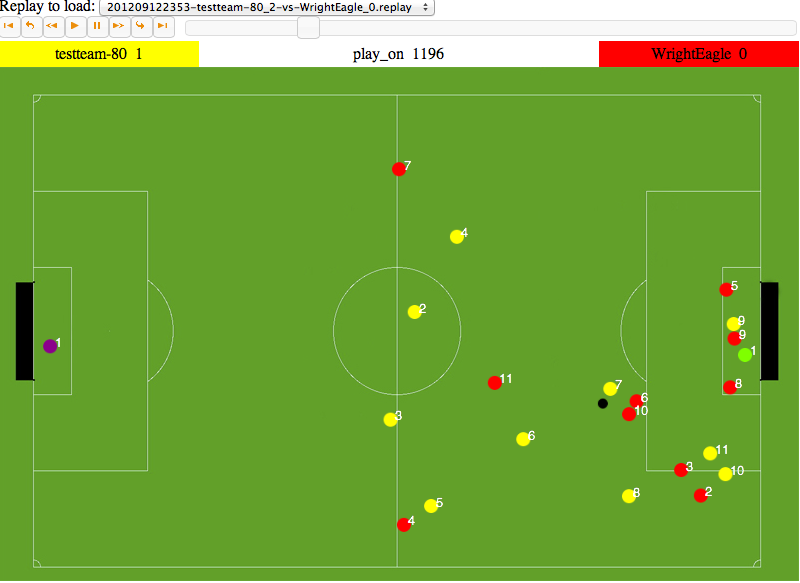}
\caption{HTML5 Logplayer developed by us. To reduce file transmission times, log files are converted into a compact textual representation, and can be selected using a HTML drop down menu.}
\label{fig:screenshot}
\end{figure}

We created a new set of programs to visualize matches in a browser window. This time, the main implementation uses HTML5 / javascript, with a pre-processing step to create simple-to-parse log files that can be conveniently transferred over the internet. The current state of the visualization basic but functional, and released by us to the RoboCup community in the hope that it will be improved over time, re-released, and be used for showing games to general public in future competitions. Figure~\ref{fig:screenshot} shows a screenshot.

The steps required to produce a HTML5 visualization from a stored log file (\texttt{*.rcg.gz}) are as follows:
\begin{enumerate}
\item Convert the log file into an intermediate text file. The required tool for this is distributed with the soccer server (\texttt{rcg2txt}).
\item Convert the intermediate text file into a \texttt{*.replay} file. This is also a text file, but with only the essential information to display the match.
\item The  \texttt{*.replay} file then can be loaded from a PHP / HTML file over a browser. The entire log player consists of a PHP / HTML script and a few javascript and CSS files.
\end{enumerate}

The log player is available (along with the necessary converter) at the Gliders2012 web page \href{http://www.oliverobst.eu/research/robotics-gliders2012-simulation-league-robocup-team}{www.oliverobst.eu/research/robotics-gliders2012-simulation-league-robocup-team}. 

\subsection{Results at RoboCup 2012}

Eventually, our team also participated in RoboCup 2012 with convincing results. The tournament was played over several rounds, where Gliders2012 proceeded to the semi-finals, resulting in a 4th place. Our detailed results are as follows:

\subsubsection{Seeding round (Round 0)}

A seeding round was played, with the 18 participating teams distributed over 4 groups of 4 or 5 teams each. The results of these groups were to be used to seed all teams more fairly for the subsequent round 1.

Gliders2012 were allocated to group C with 4 other teams: GPR-2D (Brazil), MarIik (Iran), AUT\_2D (Iran), and Riton (also Iran). Gliders2012 ended up leading the group, with one draw against MarIik, and all other matches won (Table~\ref{tab:round0}). 

\subsubsection{Round 1}

All teams were to proceed to round 1. Round 1 was played in two groups of 9 (groups E and F), with Gliders2012 in group E. Other teams in group E were: Warthog, GPR-2D, and ITAndroids (all Brazil), Oxsy (Romania), GDUT\_TiJi, YuShan2012, and WrightEagle (all China), as well as Riton (Iran).

Gliders2012 won 3 matches, drew 3, and lost 2, resulting in a 3rd place ranking in this group (Table~\ref{tab:round1}).  

\begin{table}[htdp]
\begin{minipage}[t]{0.45\linewidth}
\caption{Round 0, group C results.}
\centering
\begin{tabular}{rlcccccc}
Place & Team & Points & Total Score & W & D & L \\ \hline
1 & Gliders2012 & 10  & 9:2 & 3 & 1 & 0 \\
2 & MarIiK          &   8  & 4:2 & 2 & 2	& 0 \\
3 & GPR-2D	  &   4  & 2:3 & 1 & 1	& 2 \\
4 & AUT\_2D      &   4  & 2:3 & 1 & 1 & 2 \\
5 & Riton             &  1  & 0:7 & 0 & 1 & 3 \\
\end{tabular}
\label{tab:round0}
\end{minipage}
\hfill
\begin{minipage}[t]{0.45\linewidth}
\caption{Round 1, group E results.}
\centering
\begin{tabular}{rlcccccc}
Place & Team & Points & Total Score & W & D & L \\ \hline
1 &		WrightEagle &	24	& 53 : 7	& 8	& 0	& 0 \\
2 &		YuShan2012 & 15 & 15 : 12 & 4 & 3 & 1 \\
3 &		Gliders2012 & 12 & 18 : 17 & 3 & 3 & 2 \\
4 &		ITAndroids &11 & 13 : 19 & 3 & 2 & 3 \\
5 &		Riton	 & 10 &10 : 10 & 2 & 4 & 2 \\
6 &		GDUT\_TiJi &10 & 18 : 19 & 3 & 1 & 4 \\ \cdashline{1-7}
7 &		GPR-2D & 9 & 5 : 10 & 2 & 3 & 3 \\
8 &		Oxsy	 & 6 & 10 : 27 &  2 & 0 & 6 \\
9 &		Warthog & 2 & 4 : 25 & 0 & 2 &6 \\ 
\end{tabular}
\label{tab:round1}
\end{minipage}
\end{table}%

\begin{table}[htdp]
\begin{minipage}[t]{0.45\linewidth}
\caption{Round 2, group G results.}
\centering
\begin{tabular}{rlcccccc}
Place & Team & Points & Total Score & W & D & L \\ \hline
1 & WrightEagle & 15 & 19 : 3 & 5 & 0 & 0 \\
2 & robOTTO & 9 & 9 : 7 & 3 & 0 & 2 \\
3 & MarIiK	 & 9	& 7 : 8 & 3 & 0 & 2 \\
4 & Gliders2012 & 9	& 9 : 11 & 3 & 0 & 2 \\ \hdashline
5 & FCPortugal & 3 & 7 : 11 & 1 & 0 & 4 \\
6 & Riton & 0 & 8 : 19 & 0 & 0 & 5 \\
\end{tabular}
\label{tab:round2}
\end{minipage}
\hfill
\begin{minipage}[t]{0.45\linewidth}
\caption{Round 3, group I results.}
\centering
\begin{tabular}{rlcccccc}
Place & Team & Points & Total Score & W & D & L \\ \hline
1 & HELIOS2012 & 9 & 4 : 0 & 3 & 0 & 0 \\
2 & Gliders2012 & 6	& 5 : 1 & 2 & 0 & 1 \\ \hdashline
3 & AUT\_2D & 3	& 1 : 3 & 1 & 0 & 2 \\
4 & robOTTO & 0 & 0 : 6 & 0 & 0 & 3 \\
\end{tabular}
\label{tab:round3}
\end{minipage}
\end{table}%

\subsubsection{Round 2}
The top 6 teams of both round 1 groups proceeded to round 2, groups G and H. Gliders2012 played in group G, against Riton and MarIik (Iran), FCPortugal (Portugal), robOTTO (Germany), and WrightEagle (China). Gliders2012 won 3 matches, and lost 2, and ended up ranked 4th with an equal number of points to the second and third place. 

\subsubsection{Round 3}
The top 4 teams of both groups in round 2 proceeded to round 3, groups I and J. The group I teams were HELIOS2012 (Japan), Gliders2012, AUT\_2D (Iran), and robOTTO (Germany). With two won matches and one lost match in this group, Gliders2012 proceeded to the semi-finals.

\subsubsection{Semi-Finals and 3rd place match}
Gliders played WrightEagle for the semi-finals (2 matches), both 0:2. The subsequent match for the third place was lost against MarIiK (0:1). HELIOS2012 won the tournament, WrightEagle became runner-up.

\section{Conclusion}

We described a novel mechanism utilizing action-dependent evaluation functions, having applied it in the RoboCup Simulation 2D. The mechanism can be contrasted with some well known constructive models used by belief revision and belief update \cite{Peppas96}. The approach also allowed us to draw parallels with evaluation functions employed by chess-playing computers, in terms of mobility, field control, tropism, etc. The evaluation function that varies desirable states dependent on contemplated actions is applicable in both ball-controlling and positioning scenarios. The tactics that correspond to multiple desirable states are not imposed in a top-down fashion, but rather contribute to the evaluation via these desirable states. 

Our proposed approach and its implementation has shown to be successful both in many experiments as well as during tournament.

\newpage
\subsubsection*{Acknowledgments}
The Authors are thankful to Valentina Cupac, Andrew Curline, Tim D'Adam, Ivan Duong, James Nugent, Tom Stewart for their contribution. Team logo was created by Matthew Chadwick. Some of the Authors have been involved with RoboCup Simulation 2D in the past, however the code of their previous teams (Cyberoos and RoboLog, see, e.g., \cite{Prokopenko03,OB06}) is not used in Gliders2012.


\begin{thebibliography}{10}

\bibitem{agent2d}
Hidehisa Akiyama.
\newblock {Agent2D Base Code}.
\newblock \url{http://www.rctools.sourceforge.jp}, 2010.

\bibitem{Akiyama2008}
Hidehisa Akiyama and Itsuki Noda.
\newblock Multi-agent positioning mechanism in the dynamic environment.
\newblock In Ubbo Visser, Fernando Ribeiro, Takeshi Ohashi, and Frank Dellaert,
  editors, {\em {RoboCup 2007: Robot Soccer World Cup XI}}, pages 377--384.
  Springer, Berlin, Heidelberg, 2008.

\bibitem{Akiyama2010}
Hidehisa Akiyama and Hiroki Shimora.
\newblock Helios2010 team description.
\newblock In {\em RoboCup 2010: Robot Soccer World Cup XIV}, volume 6556 of
  {\em Lecture Notes in Computer Science}. Springer, 2011.

\bibitem{Asada99}
Minoru Asada, Hiroaki Kitano, Itsuki Noda, and Manuela Veloso.
\newblock {R}obo{C}up: {T}oday and tomorrow -- {W}hat we have have learned.
\newblock {\em Artificial Intelligence}, 110:193--214, 1999.

\bibitem{Bai2011}
Aijun Bai, Xiaoping Chen, Patrick MacAlpine, Daniel Urieli, Samuel Barrett, and
  Peter Stone.
\newblock Wrighteagle and ut austin villa: Robocup 2011 simulation league
  champions.
\newblock In {\em RoboCup 2011: Robot Soccer World Cup XV}, Lecture Notes in
  Artificial Intelligence. Springer, 2012.

\bibitem{Butler2001}
Marc Butler, Mikhail Prokopenko, and Thomas Howard.
\newblock Flexible synchronisation within {RoboCup} environment: A comparative
  analysis.
\newblock In {\em RoboCup 2000: Robot Soccer World Cup IV}, pages 119--128,
  London, UK, 2001. Springer.

\bibitem{Manual2002}
Mao Chen, Klaus Dorer, Ehsan Foroughi, Fredrick Heintz, ZhanXiang Huang, Spiros
  Kapetanakis, Kostas Kostiadis, Johan Kummeneje, Jan Murray, Itsuki Noda,
  Oliver Obst, Pat Riley, Timo Steffens, Yi~Wang, and Xiang Yin.
\newblock {\em Users Manual: {R}obo{C}up Soccer Server --- for Soccer Server
  Version 7.07 and Later}.
\newblock The {R}obo{C}up Federation, February 2003.

\bibitem{DFL08}
Frank Dylla, Alexander Ferrein, Gerhard Lakemeyer, Jan Murray, Oliver Obst,
  Thomas R{\"o}fer, Stefan Schiffer, Frieder Stolzenburg, Ubbo Visser, and
  Thomas Wagner.
\newblock {\em Computers in Sport}, chapter Approaching a Formal Soccer Theory
  from the Behavior Specification in Robotic Soccer, pages 161--186.
\newblock Bioengineering. WIT Press, 2008.

\bibitem{GO03}
Thilo Girmann and Oliver Obst.
\newblock robocup2flash version 0.3.
\newblock
  \url{http://robolog.cvs.sourceforge.net/viewvc/robolog/robocup2flash/}, 2003.

\bibitem{Kitano97}
Hiroaki Kitano, Milind Tambe, Peter Stone, Manuela~M. Veloso, Silvia
  Coradeschi, Eiichi Osawa, Hitoshi Matsubara, Itsuki Noda, and Minoru Asada.
\newblock {The RoboCup Synthetic Agent Challenge 97}.
\newblock In {\em RoboCup-97: Robot Soccer World Cup I}, pages 62--73, London,
  UK, 1998. Springer.

\bibitem{Chess00}
Fran\c{c}ois~Dominic Laram{\'e}e.
\newblock {Chess Programming Part VI: Evaluation Functions}.
\newblock
  \url{http://www.gamedev.net/page/resources/\_/technical/artificial-intelligence/chess-programming-part-vi-evaluation-functions-r1208},
  2000.

\bibitem{Noda2003}
Itsuki Noda and Peter Stone.
\newblock {The RoboCup Soccer Server and CMUnited Clients: Implemented
  Infrastructure for MAS Research}.
\newblock {\em Autonomous Agents and Multi-Agent Systems}, 7(1--2):101--120,
  July--September 2003.

\bibitem{OB06}
Oliver Obst and Joschka Boedecker.
\newblock Flexible coordination of multiagent team behavior using {HTN}
  planning.
\newblock In Itsuki Noda, Adam Jacoff, Ansgar Bredenfeld, and Yasutake
  Takahashi, editors, {\em {R}obo{C}up 2005: Robot Soccer World Cup~{IX}},
  Lecture Notes in Artificial Intelligence, pages 521--528. Springer, Berlin,
  Heidelberg, New York, 2006.

\bibitem{Peppas96}
Pavlos Peppas, Abhaya~C. Nayak, Maurice Pagnucco, Norman~Y. Foo, Rex Bing~Hung
  Kwok, and Mikhail Prokopenko.
\newblock Revision vs. update: Taking a closer look.
\newblock In Wolfgang Wahlster, editor, {\em 12th European Conference on
  Artificial Intelligence, Budapest, Hungary, August 11-16, 1996, Proceedings},
  pages 95--99. John Wiley and Sons, Chichester, 1996.

\bibitem{POWH12}
Mikhail Prokopenko, Oliver Obst, Peter Wang, and Jason Held.
\newblock Gliders2012: Tactics with action-dependent evaluation functions.
\newblock In {\em {R}obo{C}up 2012: Team Description Paper}, Mexico City,
  Mexico, June 2012.

\bibitem{Prokopenko02}
Mikhail Prokopenko and Peter Wang.
\newblock Relating the entropy of joint beliefs to multi-agent coordination.
\newblock In Gal~A. Kaminka, Pedro~U. Lima, and Ra{\'u}l Rojas, editors, {\em
  RoboCup 2002: Robot Soccer World Cup VI}, volume 2752 of {\em Lecture Notes
  in Computer Science}, pages 367--374. Springer, 2003.

\bibitem{Prokopenko03}
Mikhail Prokopenko and Peter Wang.
\newblock Evaluating team performance at the edge of chaos.
\newblock In Daniel Polani, Brett Browning, Andrea Bonarini, and Kazuo Yoshida,
  editors, {\em RoboCup 2003: Robot Soccer World Cup VII}, volume 3020 of {\em
  Lecture Notes in Computer Science}, pages 89--101. Springer, 2004.

\bibitem{Reis2001}
Lu\'{\i}s~Paulo Reis, Nuno Lau, and Eugenio Oliveira.
\newblock Situation based strategic positioning for coordinating a team of
  homogeneous agents.
\newblock In {\em Balancing Reactivity and Social Deliberation in Multi-Agent
  Systems, From RoboCup to Real-World Applications (selected papers from the
  ECAI 2000 Workshop and additional contributions)}, pages 175--197, London,
  UK, 2001. Springer.

\bibitem{ATAL2000}
Patrick Riley, Peter Stone, and Manuela Veloso.
\newblock Layered disclosure: Revealing agents' internals.
\newblock In C.~Castelfranchi and Y.~Lesperance, editors, {\em Intelligent
  Agents VII. Agent Theories, Architectures, and Languages ---
  7th.~International Workshop, ATAL-2000, Boston, MA, USA, July 7--9, 2000,
  Proceedings}, Lecture Notes in Artificial Intelligence. Springer, Berlin,
  Berlin, 2001.

\bibitem{IAAI2000}
Peter Stone, Patrick Riley, and Manuela Veloso.
\newblock Defining and using ideal teammate and opponent models.
\newblock In {\em Proceedings of the Twelfth Annual Conference on Innovative
  Applications of Artificial Intelligence}, 2000.

\bibitem{Stone99}
Peter Stone and Manuela Veloso.
\newblock Task decomposition, dynamic role assignment, and low-bandwidth
  communication for real-time strategic teamwork.
\newblock {\em Artificial Intelligence}, 110(2):241--273, June 1999.

\bibitem{Stone98}
Peter Stone and Manuela~M. Veloso.
\newblock Team-partitioned, opaque-transition reinforced learning.
\newblock In {\em RoboCup-98: Robot Soccer World Cup II}, pages 261--272,
  London, UK, 1999. Springer.

\end{thebibliography}
\end{document}